\documentclass[11pt]{article}
\usepackage[margin=1in]{geometry}
\usepackage{amsmath}
\usepackage{amsfonts}
\usepackage{amssymb}
\usepackage{tikz-cd}
\usepackage{enumitem}
\usepackage{booktabs}
\usepackage{tabularx}
\usepackage{array}
\usepackage{longtable}
\usepackage{tcolorbox}
\usepackage{amsthm}
\usepackage[square,comma,authoryear]{natbib}

\newtheorem{lemma}{Lemma}
\newtheorem{corollary}{Corollary}
\newtheorem{proposition}{Proposition}
\newtheorem{definition}{Definition}
\newtheorem{remark}{Remark}
\newtheorem{example}{Example}

\title{Towards Error-Centric Intelligence I:\\Beyond Observational Learning}
\author{Marcus A. Thomas \thanks{This work was conducted independently and does not represent the views of Memorial Sloan Kettering.} \\\texttt{thomm15@mskcc.org}}
\date{\today}

\begin{document}

\maketitle
\begin{abstract}
We argue that progress toward AGI is theory-limited rather than data- or scale-limited. Building on Deutsch–Popper critical rationalism, we challenge the Platonic Representation Hypothesis: observationally equivalent worlds can diverge under interventions, so observational adequacy alone cannot guarantee interventional competence. We begin by laying foundations—definitions of knowledge, learning, intelligence, counterfactual competence, and AGI—and then analyze the limits of observational learning that motivate an error-centric shift. We recast the problem as three questions about (i) how explicit and implicit errors evolve under an agent’s actions, (ii) which errors are unreachable within a fixed hypothesis space, and (iii) how conjecture and criticism expand that space.

From these questions we propose \emph{Causal Mechanics}, a mechanisms-first program in which hypothesis-space change is a first-class operation and probabilistic structure is used when useful rather than presumed. We advance structural principles that make error discovery and correction tractable: a differential Locality–Autonomy Principle (LAP) for modular interventions, a gauge-invariant form of Independent Causal Mechanisms (ICM) for separability, and the Compositional Autonomy Principle (CAP) for analogy preservation, together with actionable diagnostics. The aim is a scaffold for systems that can convert unreachable errors into reachable ones and correct them.
\end{abstract}

\section{Introduction}
\label{sec:intro}

Many in the AI research community believe the path to artificial general intelligence (AGI) to be data limited\textemdash including limitations in model capacity and task variety. In this perspective, the fundamental breakthroughs, large neural network architectures fed by large datasets, have been made. Maybe they would admit that a few new ideas are needed to achieve planning, autonomy, causal reasoning, etc., but these are fundamentally unimportant compared with the scaling laws we have already discovered and will continue to discover via engineering progress. The prediction made is that larger and better datasets will lead to smarter and more capable learning systems.

We argue that this data-driven paradigm is wrong, that AGI is fundamentally theory limited. We assume that (i) humans exhibit general intelligence and (ii) other instances of general intelligence\textemdash biological or artificial, terrestrial or otherwise\textemdash are in principle possible. We also argue that the defining capabilities of AGI must be expressible without explicit reference to current human tasks. Prehistoric Homo sapiens certainly possessed general intelligence, and closely related extinct human species probably did as well \citep{Kozowyk2017BirchTarSciRep, Schmidt2023KoenigsaueAAS, Jaubert2016BruniquelNature, Hardy2020CordageSciRep, Hoffmann2018CaveArtScience, Pomeroy2020ShanidarAntiquity}.

Our goal is not to propose an explanatory theory of AGI\footnote{Such a theory might explain the relationship between general intelligence and consciousness and make testable predictions.}, but to provide definitions, formalism, and arguments that apply to biological and non-biological systems and may be useful to a future theory.

\noindent\textbf{Overview.}
Section~\ref{sec:intro} relates representations, hypotheses, knowledge, learning, and Systems~1/2, culminating in a formal definition of AGI. Section~\ref{sec:platonic} analyzes limits of observational learning and the Platonic Representation Hypothesis. Section~\ref{sec:error_questions} reframes general intelligence as three error–centric questions on error evolution, representational reach, and conjecture–criticism capacity. Section~\ref{sec:principles} states the resulting structural commitments—LAP for modular interventions, a gauge–invariant ICM for separability, and CAP for analogy preservation—together with diagnostic witnesses. Section~\ref{sec:conclusion} concludes and points to Part~II (E–SCMs) for a modeling approach.

\subsection{Knowledge without Learning}

Why do we all agree that simple pocket calculators are not instances of AGI? The calculator's internal state certainly contains knowledge relevant to performing certain computations. In fact, we can conceptualize its computations in the same terms we use to analyze deep learning systems. The calculator can be thought of as a two-stage mapping $h = g \circ f$ applied each time you press ``=''. The encoder $f$ embeds user input into a representation space that is suitable for the fixed set of functions the device can perform. The head $g$ is the fixed computational circuitry that maps such an embedding to the appropriate result which is then displayed. The precise $f$/$g$ split is really a modeling abstraction, not necessarily a claim about separate hardware blocks or computing modules.

For example, we may adopt a single-hypothesis view in which the calculator implements a unified computational pipeline and realizes exactly one hypothesis $h = g \circ f$, making the hypothesis space a singleton: $\mathcal{H} = \{g \circ f\}$. Alternatively, in a multi-hypothesis view we might decompose the calculator more finely, e.g., the encoder $f$ parses the keystroke sequence into a more structured mathematical representation, perhaps a parse tree or sequence of (operand, operator) pairs that preserves precedence and associativity. Different heads could represent different evaluation strategies: $g_{\text{standard}}$ applies standard order of operations (PEMDAS), $g_{\text{left-to-right}}$ evaluates strictly left-to-right ignoring precedence, $g_{\text{safe}}$ adds overflow checking, etc. Each strategy $g_t$ processes the same structured representation from $f$ but implements different computational policies. The hypothesis space becomes:
\[
\mathcal{H} = \left\{g_{\text{standard}} \circ f, g_{\text{left-to-right}} \circ f, g_{\text{safe}} \circ f, \ldots\right\}.
\]

An important implication is that different hypothesis spaces for a system entail distinct error-diversity profiles, discovery mechanisms, and correction strategies. Any system designer (or evolutionary process shaping functionality) must navigate these inherent error\textendash representation trade-offs. 

\subsection{From Data-Centric to Error-Centric Intelligence}
This shows that sophisticated error structures alone do not constitute intelligence without mechanisms to transform discovered errors into knowledge, that is, to learn.

\paragraph{Closed hypothesis by design.} Modern LLM systems are analogous to the calculator in the sense that their hypothesis spaces are also static.
For a fixed architecture $\mathcal{A}$ (block layout, attention, activations, normalization), tokenizer and output alphabet $\mathcal{V}$, and context interface with maximum length $L$, the model realizes a family of conditional distributions
\[
\mathcal{H}(\mathcal{A},\mathcal{V},L)=\big\{\,p_\theta(\cdot\mid x_{<t}) : \theta\in\Theta(\mathcal{A})\,\big\}.
\]
Training procedures—pretraining, supervised fine-tuning, and RLHF—move $\theta$ within this family but do not alter its boundary. Decoding and test-time compute only change how samples are drawn from $p_\theta$; they do not enlarge $\mathcal{H}$. The hypothesis class itself changes only under interface or architectural edits (e.g., a new tokenizer, longer $L$, added modalities, external memory, or mechanism-level modules). Absent explicit self-modification machinery, such edits are external engineering interventions, not consequences of the system’s own explanatory knowledge. This design fact explains why the usual training moves cannot convert unreachable errors into reachable ones: they do not add representational options, they reweight existing ones.

Modern LLM-based systems therefore learn in two limited respects. First, learning occurs only during training phases when errors durably affect internal state; at inference, errors typically do not alter $\theta$. Second, learning is confined to narrow error families such as next-token prediction or a reward model’s surrogate. In next-token prediction,
\[
\mathcal{L}_{\mathrm{NTP}}(\theta)
= \mathbb{E}_{x\sim \mathcal{D}}\sum_{t=1}^{|x|}
-\log p_\theta\!\left(x_t \mid x_{<t}\right),
\]
and gradients arise solely from token discrepancies. Supervised fine-tuning optimizes the same loss on curated pairs, while RLHF maximizes a learned reward under a KL constraint,
\[
\mathcal{L}_{\mathrm{RLHF}}(\theta)
= - \mathbb{E}_{x,\,y\sim \pi_\theta(\cdot\mid x)} \big[ R_\phi(x,y) \big]
\;+\; \beta\, \mathrm{KL}\!\big(\pi_\theta(\cdot\mid x)\,\|\,\pi_{\mathrm{ref}}(\cdot\mid x)\big),
\]
but these, too, move only within $\mathcal{H}(\mathcal{A},\mathcal{V},L)$.

These limitations can be reframed in terms of the System 1 vs System 2 distinction proposed by Kahneman \cite{kahneman2011thinking}. In current AI systems, there is a disconnect between learning (`training' in the dominant AI paradigm) and thinking or reasoning. We conjecture that what separates Systems 1 and 2 is the degree to which they create explanatory knowledge. The development of AGI systems, which are capable of System 2 operation in this sense, requires asking fundamental questions about error discovery, knowledge formation, representational reach, and the capacity for conjecture and criticism.

\subsection{Foundational Definitions}
The definitions in this section are proposed conventions, terminological choices intended to scaffold inquiry into general intelligence and AGI. They are not empirical claims but are motivated by prior empirical and philosophical work. Falsifiable content appears elsewhere in the text, including our theorems, propositions, diagnostics, etc.

\begin{definition}[Knowledge and Explanatory Knowledge]\label{def:knowledge}
\footnotesize\emph{Attribution.} This formulation is inspired by David Deutsch and Chiara Marletto's constructor-theoretic account of information and knowledge (see, e.g., \citep{deutsch2013constructor, deutsch2015constructor,deutsch2011beginning}).\normalsize

Knowledge is information that causally contributes to its own persistence via copying or retention. Explanatory knowledge is the subset of knowledge that (i) is produced and maintained in the course of problem solving by a cycle of conjecture, criticism, and error correction; (ii) supports counterfactual and interventional reasoning (e.g., abduction\textendash intervention\textendash prediction coherence); and (iii) accounts for observations by positing mechanisms over unobserved reality. 
\end{definition}

\begin{definition}[Learning]\label{def:learning}
Learning is the process by which a system uses criticism signals\textemdash such as errors in prediction, failures in goal attainment, violations of constraints, and inconsistencies revealed by reasoning or limitations in representation\textemdash to cause knowledge to become durably embedded in its internal state.
\end{definition}
Explanatory knowledge and learning enable open-ended problem solving.
\begin{definition}[Intelligence]\label{def:intelligence}
Intelligence is any measure of the  efficiency with which a system creates explanatory knowledge. 
\end{definition}
Creation of explanatory knowledge includes both the invention of new explanatory structures and the improvement of existing ones via error correction. It includes refinement, replacement, and synthesis of prior knowledge insofar as these revisions introduce new explanatory structure or expand the system’s representational or interventional reach. Synthesis counts as creation only when it yields new counterfactual commitments; otherwise it is transformation without epistemic gain. See Appendix \ref{sec:appendix-IntelOp} for a discussion of intelligence measures.
\begin{definition}[Competence]\label{def:competence}
Competence is any measure of the ability to use explanatory knowledge to solve problems. 
\end{definition}

Based on these definitions, the processes of evolution by natural selection can be understood as creating knowledge (e.g., knowledge for survival and reproduction in an environment which is embedded in DNA), but there is no operative intelligence because the knowledge is not explanatory. It also follows that many modern AI systems possess high competence but low intelligence, even though they can learn vast quantities of explanatory knowledge via training.

\begin{definition}[Counterfactual Competence and Understanding]\label{def:understanding}
An agent has counterfactual competence when it can (i) represent mechanism-changing hypotheses, (ii) manipulate them via model surgery that specifies which mechanisms change and which remain invariant, and (iii) learn by exploring the implications of counterfactuals. Understanding is the capability of using counterfactual competence to generate explanations (interdependent conjectures exhibiting hard-to-vary structure). The degree of understanding increases with the degree of hard-to-vary structure. 
\end{definition}

\begin{definition}[Artificial General Intelligence]\label{def:agi}
An AGI is a non-biological general intelligence, that is, a system capable of the unbounded creation and improvement of explanatory knowledge. 
\end{definition}
Unbounded\footnote{Constrained only by the laws of physics.} explanatory knowledge creation requires open-ended error discovery, unbounded error correction, and the ability to learn. Constrained agency (\textit{policy following}; the capacity to select and execute actions to satisfy a specified policy or objective set) and autonomous agency
(\textit{policy authoring}; the capacity to use explanatory knowledge acquired through learning to create, modify, or delete one's own policies and objectives) are requirements of the qualifiers ``open-ended" and ``unbounded". Such a system can improve its understanding indefinitely.

Based on this definition of AGI, an implication is that such a system can in principle create the same explanatory knowledge as a human. However, it does not follow that all tasks or domains can be learned with the same efficiency or reliability. Human brains evolved via natural selection, and therefore our strengths and weaknesses may differ substantially from those of designed systems. What unites general intelligences is their generality rather than their degree of intelligence across domains. Just as two universal Turing machines can emulate one another (ignoring efficiency), any two AGIs can, in principle, recreate each other's explanatory knowledge. The reason is that structural, open-ended error discovery plus unbounded error correction (with learning and autonomy) allows each to effectively reproduce equivalent conjectures, tests, policies, and ultimately, surviving theories. 

It is important not to conflate definitions with assays, the procedures for quantifying some of the possible implications of the definition. Although no tests\footnote{E.g., benchmarks, the ARC AGI program, expectations that AGI implies x\% growth rate in or market share of the economy, etc.} can be definitive, a portfolio of tests designed in light of a theory of how a particular putative AGI works may better triangulate the construct while resisting metric gaming.

\begin{definition}[Synthetic conjecture]
A synthetic conjecture is any representational commitment whose content is not logically entailed by observations, nor by deductive consequences of an already adopted framework, yet carries testable consequences. Such conjectures are made relative to a problem–situation (e.g., an explanatory or design task), even when the problem is only partially specified. 
\end{definition}
Examples include the adoption or redesign of a hypothesis space or model class, the choice of priors, codes, or minimum description length penalties, the introduction of new operators or semantics, and the assertion of representability by a structural causal model.
The term `synthetic' is used in the Kantian–Popperian sense of ampliative: such conjectures extend what is given by observation. The intervention calculus is a canonical example: positing intervention semantics (e.g., realizing $\mathrm{do}(X{=}x)$) adds structure not derivable from observational distributions; deciding which concrete manipulation is represented as $\mathrm{do}(X{=}x)$—and which invariances the surgery assumes—is itself a synthetic conjecture.

\subsection{Epistemological Foundations}

\begin{tcolorbox}[
    colback=gray!5,
    colframe=gray!50,
    title={\textbf{Core Epistemological Critiques of Inductive Learning}},
    fonttitle=\sffamily,
    boxrule=0.5pt,
    arc=2pt,
    left=6pt,
    right=6pt,
    top=4pt,
    bottom=4pt
]

\noindent
\textbf{Pearl (intervention $\neq$ conditioning).}
Interventions, formalized with $\mathrm{do}(\cdot)$ \citep{pearl2000models,pearl2018book}, are defined by model surgery: replacing the mechanism for a target variable while holding others invariant. This is not a Bayesian conditioning move within a fixed hypothesis space; it is a change to the hypothesis space itself (a structural alteration of the model’s mechanisms). Counterfactual competence implicitly depends on this notion of intervention.

\vspace{3pt}
\noindent
\textbf{Popper (conjecture and refutation).}
Knowledge grows by proposing non-derivable\footnote{The explanatory content of a new hypothesis is not logically entailed by observational statements  \citep{popper1959logic,popper1963conjectures}} conjectures and subjecting them to severe tests. 

\vspace{3pt}
\noindent
\textbf{Popper--Miller (no inductive support from probability-raising).}
Within probability theory, raising the probability of a hypothesis by conditioning on favorable evidence does not supply the missing explanatory content of that hypothesis. Apparent ``inductive support'' is either deductive (arising from logical containment) or a redistribution of credence across already-specified possibilities; it does not originate new universal or counterfactual structure \citep{popper1983proof}.

\vspace{3pt}
\noindent
\textbf{Deutsch (explanatory knowledge).}
Understanding requires explanations—accounts of what is seen in terms of unseen mechanisms—whose content is hard to vary while still accounting for the phenomena \citep{deutsch2011beginning}.

\vspace{3pt}
\noindent
\textbf{Our Synthesis (synthetic conjecture $\rightarrow$ changes hypothesis space $\rightarrow$ enables error discovery).}
Unbounded error discovery requires the ability to change the hypothesis space itself—to introduce or remove variables, propose new mechanisms, alter intervention semantics (e.g., the rules by which you interpret $\mathrm{do}(\cdot)$) and refactor invariances (which relations are universal versus context-specific).   
\end{tcolorbox}

\section{Limitations of Observational Learning}\label{sec:platonic}

The dominant paradigm in machine learning embodies a fundamentally inductive\footnote{For example, extrapolation, interpolation, and imputation from finite observations to claims about unseen reality.} chain:
\begin{equation}
P_{\rm obs} \;\xrightarrow{\text{data collection}}\; \hat{P}_{\rm obs} \;\xrightarrow{\text{ERM}}\; f^* \;\xrightarrow{\text{convergence}}\; \text{``any downstream task."}
\end{equation}
This assumes passive observation suffices for representation learning, understanding, and open-ended problem solving, irrespective of the relevance of causal identifiability to the task.\footnote{A causal effect is identifiable iff its estimand can be rewritten entirely free of the do-operator. Without causal assumptions, one cannot distinguish whether $X \rightarrow Y$, $X \leftarrow Y$, or $X \leftarrow C \rightarrow Y$\textemdash all can produce the same correlations but yield different $P(Y \mid \mathrm{do}(X))$.} As a result, entire classes of implicit errors\textemdash those living in the gap between $P_{\rm obs}$ and interventional distributions\textemdash remain unreachable until they manifest as explicit failures. 

\subsection{The Platonic Representation Hypothesis}

We analyze the claim that large-scale observational learning yields a single, universal latent geometry. For clarity, we fix notation: $P_{\mathrm{obs}}$ denotes the observational law over data; $P_{\text{reality}}$ names the hypothesized ``shared model of reality" in the platonic view. Encoders $f$ summarize observations; heads $g$ answer queries from those summaries\footnote{For transformer architectures (encoder-only, decoder-only, encoder–decoder; including LLMs), there is no single fixed “one-time” latent $f(x)$: token states are updated across layers by self-attention and feed-forward blocks, so representations are context-dependent. The $f/g$ split is a modeling convenience: one may treat the stack up to the final output projection as $f$ and the projection/softmax as $g$, but this is not a hard architectural boundary.}.

Modern scaling practice implicitly posits a stationary joint $P_{\text{reality}}(\mathcal Z)$. Learning then reduces to making $P_{\mathrm{obs}}$ approach $P_{\text{reality}}$. Under this observational view, sending samples through $f$ yields the push\textendash forward
\[
  P_{\text{reality}}^{f}(B) \;=\; P_{x\sim P_{\text{reality}}}\bigl(f(x)\in B\bigr).
\]
Fitting an encoder\textendash head pair $h=g\circ f$ minimizes empirical risk
\[
\widehat R(h)=\mathbb E_{x\sim P_{\mathrm{obs}}} L\!\bigl(h(x),t(x)\bigr),
\quad \text{aiming at } R(h)=\mathbb E_{x\sim P_{\text{reality}}} L\!\bigl(h(x),t(x)\bigr),
\]
with uniform\textendash convergence controlling the gap on that same observational slice.\footnote{We keep ``observational slice" explicit because nothing here guarantees stability beyond the distribution that generated the observations.}

The premise is articulated in \citep{huh2024platonic} as a shared statistical model of reality. The Platonic Representation Hypothesis (PRH) concerns the geometry of vector embeddings—formally, functions $f\!:\!X\!\to\!\mathbb{R}^n$ whose induced kernels measure similarity among datapoints—and argues that, as models and data scale, these geometries increasingly align across architectures and modalities, converging toward a common latent structure, a `platonic representation'\citep{huh2024platonic}. In our view, the available evidence is pipeline-conditional, a product of the standard observational ERM + SGD training setup, rather than an insight into reality itself. If training were done under explicit causal interventions or invariance constraints, convergence claims could differ.

\subsubsection*{Implications of a strict interpretation}

Under a strict interpretation, the platonic representation produced by encoder $f$ alone would capture all necessary structure so that any head $g$ could answer interventional and counterfactual queries. This fails because different causal data-generating processes can induce the same $P_{\mathrm{obs}}(X,Y)$ yet yield different $P(Y\mid \mathrm{do}(X{=}x))$; hence, absent additional causal assumptions (e.g., conditions ensuring identifiability), no functional $F\!\left(P_{\mathrm{obs}},x\right)$ recovers interventional responses in general.

\begin{proposition}[Standard, after \cite{pearl2000models}]
\label{thm:obs-not-enough}
Purely observational data do not, in general, identify interventional laws when causal structures are observationally equivalent.\footnote{Construct distinct SCMs (e.g., $X\!\to\!Y$, $Y\!\to\!X$, and $X\!\leftarrow\!C\!\to\!Y$)
that induce the same observational joint $P_{\mathrm{obs}}(X,Y)$ but yield different
$P\!\left(Y\mid \mathrm{do}(X{=}x)\right)$. Any $F$ that depends only on $P_{\mathrm{obs}}$ would have to
output two different numbers on the same input, a contradiction.}
\end{proposition}
This observation aligns with results on the causal–neural connection showing that universal function approximators do not bypass identifiability limits: even arbitrarily expressive neural models cannot, in general, recover interventional laws from observational data alone \citep{xia2021causal}.

\subsubsection*{A relaxed interpretation: causal-aware heads}

A relaxed view concedes that the platonic representations themselves need not encode causal structure; instead, task heads $g$ supply it. The composite $h=g\circ f$ can answer interventional queries only if $g$ brings prior causal assumptions (architecture or constraints). However, joint training of $f$ with $g$ would likely pressure $f$ to retain features that distinguish causally distinct but observationally equivalent worlds, which may undermine any unique platonic geometry.

\subsection{Catastrophic Forgetting and the Fractured–Entangled Representation Hypothesis}

Catastrophic forgetting\textendash the overwriting of earlier competence by later training\textendash need not be viewed as an incidental stability–plasticity failure. We frame it instead as a structural consequence of fractured, entangled representations (FER) \citep{kumar2025questioning} produced by conventional observational learning via SGD: fracture when information underlying the same unitary concept is split into disconnected, redundant pieces, and entanglement when those fractured functions inappropriately influence one another rather than remaining modular. By extension, in task head
$g$, the same pathologies manifest themselves at the level of task mappings: fracture when a single query type is redundantly implemented across disjoint fragments of the head, and entanglement when distinct queries share overlapping output circuitry. Both levels amplify interference risk in sequential training, making catastrophic forgetting the temporal face of fractured, entangled design. The combination makes interference the default: any gradient that acquires a new capability is liable to traverse parameters that also realize old ones. The contrasting case is a unified, factored representation (UFR): each capability is encoded once and factorized from others, so that learning can be localized and composable without collateral damage.

Rather than treating catastrophic forgetting as an optimization glitch to be patched\footnote{Attempts span four broad families. (1) \emph{Replay} reuses past data (or pseudo-data) during new learning: exemplar buffers and gradient-projection variants (e.g., GEM/A-GEM), strong baselines like ER/DER, and language-model–style generative replay (DGR, LAMOL) \citep{LopezPaz2017GEM,Chaudhry2019AGEM,Buzzega2020DER,Shin2017DGR,Sun2019LAMOL}. (2) \emph{Regularization/distillation} constrains parameter drift or output drift to preserve prior functions: EWC/Fisher penalties, Synaptic Intelligence, Memory-Aware Synapses, and Learning-without-Forgetting \citep{Kirkpatrick2017EWC,Zenke2017SI,Aljundi2018MAS,Li2016LwF}. (3) \emph{Parameter isolation / expansion} preserves old skills by allocating task-scoped routes or weights: Progressive Nets, PackNet pruning-and-freezing, Piggyback masks, adapters/AdapterFusion, and LoRA \citep{Rusu2016PNN,Mallya2018PackNet,Mallya2018Piggyback,Pfeiffer2021AdapterFusion,Hu2021LoRA}. (4) \emph{Interference control} reduces destructive gradient interactions without (much) replay: orthogonal/constrained updates and related projection schemes \citep{Farajtabar2020OGD}. Class-incremental protocols often combine these ingredients with balanced classifiers and exemplars (e.g., iCaRL) \citep{Rebuffi2017iCaRL}.}, we treat it as a representational design failure. The remedy is causality-aware learning in both \(f\) and \(g\), governed by an explicit working hypothesis \(h\) which is the subject of criticism. Framed this way, established patches can be reinterpreted as partial moves toward UFRs: replay re-imposes older gradients to counteract entangled edits; regularization penalizes drift along previously used directions; parameter isolation supplies mechanism-keyed routes ex post; orthogonalization constrains updates to live in subspaces that reduce coupling. Our proposal for causality-aware learning of \(f,g,h\) aims to achieve the same end by construction.

Interestingly, the FER result showing that an ERM+SGD model and an open-endedly evolved model can implement the same input–output mapping while realizing markedly different internal geometries either (i) contradicts a pipeline-invariant reading of PRH, or (ii) leaves a pipeline-conditional reading untouched while highlighting that any observed geometric convergence is an inductive bias of the ERM+SGD pipeline, not a universal property of the data-generating process.

\subsection{Epistemic Inadequacy}
From a critical-rationalist perspective, purely inductive accounts of learning are philosophically inadequate: knowledge advances by conjecture and criticism rather than by justifying generalizations from data. Bayesian epistemology is a leading formal framework for inference under uncertainty and belief revision, especially when the underlying mechanisms and their interactions are only partially known. In this regard, we, and possibly future AGIs, use the probability calculus, Bayes’ rule, and an explicit space of hypotheses to represent and revise degrees of belief as evidence accumulates. However, this epistemic program faces serious challenges, including the Popper–Miller result\citep{popper1983proof}, which argues that probabilistic support does not provide a genuinely inductive justification for universal hypotheses.

To be clear, by induction we mean the purported ampliative move whereby observations supply new support to the  unentailed (not logically implied by the evidence) content of a hypothesis (e.g., projecting from observed instances to universal claims). The Popper--Miller analysis denies that Bayesian conditionalization achieves this. When $P(H\mid E)>P(H)$, the increase is exhaustively accounted for by (i) deductive overlap with what $E$ already entails and (ii) at best, no positive support for the remainder of $H$ not entailed by $E$ (and often a decrease). No genuinely ampliative support accrues to unentailed content. Thus, conditionalization is a coherence-preserving, deductive re-weighting within a prior framework, not an engine for generating explanatory content.

\begin{remark}[Popper--Miller decomposition]
For any events $H,E$ with $0<P(E)<1$,
\[
P(H\mid E)-P(H)
= \underbrace{\frac{P(H\wedge E)\,P(\lnot E)}{P(E)}}_{\text{deductive overlap}}
\;-\;
\underbrace{P(H\wedge \lnot E)}_{\text{countersupport to unentailed content}}.
\]
Thus any posterior increase splits into support for the part of $H$ deductively shared with $E$, minus support for the rest; the ``increase'' is not ampliative in Popper's sense.
\end{remark}

Two standard replies deserve brief acknowledgment. First, many Bayesians care about comparative support (Bayes factors over a discrete hypothesis set). Our point does not forbid model comparison; it questions whether observational updating alone furnishes ampliative support to unconstrained content. Second, worries about zero prior mass on universals depend on representation and measure choices. Even granting the merit of these replies, the core gap remains: observational updating lacks an explicit calculus of interventions and counterfactuals. 

\paragraph{Bayes cannot express nor retrospectively identify do-operators.}

Bayesian conditionalization reweights beliefs within a fixed observational model. It neither defines do-operators nor identifies the correct one from observational data. 

\begin{proposition}[Do-operators are not Bayesian updates]
\label{prop:do-not-update}
Let $\mathcal M=(\mathcal H,\pi,\{p(\cdot\mid h)\}_{h\in\mathcal H})$ be an observational Bayesian model and let
$\mathsf{Cond}\!: (x_{1:n}\mapsto \pi(\cdot\mid x_{1:n}))$ denote Bayesian conditionalization. There is no functional
of $(\pi,\{p(\cdot\mid h)\},\mathsf{Cond})$ that yields interventional quantities
$p(y\mid \mathrm{do}(a),x_{1:n})$ \emph{without} supplying, as extra structure, a family of surgery maps
$\{\tau_a\}$ that define interventional kernels $p_a(\cdot\mid h)=\tau_a[p(\cdot\mid h)]$.
\end{proposition}

\begin{proposition}[Bayes cannot retrospectively identify do-structure]
\label{prop:do-not-identify}
Fix an observational model $\mathcal M$ as above and two surgery families $\{\tau_a\}$, $\{\tilde\tau_a\}$ that
\emph{agree on the observational regime} (they induce the same $p(\cdot\mid h)$ for all $h$). Then for any
observational dataset $x_{1:n}$,
\[
\pi(h\mid x_{1:n})=\tilde\pi(h\mid x_{1:n})
\quad\text{while}\quad
p_a(\cdot\mid h)\ \text{and}\ \tilde p_a(\cdot\mid h)\ \text{may differ.}
\]
Hence conditionalization on observational data cannot select the “correct’’ do-operator among such competitors.
\end{proposition}

In short, Bayes reweights; $\mathrm{do}(\cdot)$ rewires. Reweighting cannot express rewiring and, given multiple rewiring schemes that coincide observationally, Bayesian updating is not a general method for identifying the right one. 

Real inquiry is open. Investigators introduce novel variables, mechanisms, and model structures. The act of positing a genuinely new hypothesis and assigning it a prior weight and a likelihood function is a itself creative move, not an inference from observations. 

Once a hypothesis has been specified, Bayesian conditionalization provides a disciplined comparative audit, but this form of criticism is not exclusive. Hypotheses can be stressed via likelihood-only comparisons, frequentist severity and goodness-of-fit tests, predictive checks (prequential analyses and proper scoring rules), information and complexity penalties (AIC/BIC/MDL), causal-constraint tests across environments, and robustness/sensitivity analyses. None of these procedures, Bayesian or otherwise, generate explanatory content; they only test what creativity supplies.

The pragmatic upshot is a division of labor: (i) Creative conjecture: the introduction of a non-derivable explanatory structure that potentially includes its probabilistic scaffolding; (ii) Criticism: the comparative deductive audit (Bayes factors, predictive scores, severity tests, and the like); and (iii) Revision: refactoring the representational space in light of failures (model surgery, new variables, altered mechanisms). 

Even Invariant Risk Minimization (IRM)---which seeks to verify pre-specified invariance conjectures across predetermined environments rather than generate new hypotheses about causal structure---inherits the limits of a hypothesis-closed setting. The method optimizes
\[
\min_{f,g} \sum_e R_e(g\circ f)
\quad\text{subject to $g$ being optimal on each environment separately,}
\]
while assuming that the training environments contain sufficient diversity for invariance discovery and that the relevant invariant relationships are already specified in the objective.

As a critic, IRM can be genuinely useful: when the environments are sufficiently diverse, failures to satisfy the invariance constraints constitute informative falsifications, and performance on held-out environments can localize misspecification in $f$ or in the stated invariances. However, the system remains epistemologically static: it can detect violations of its stated invariances but cannot exploit those errors to refine the invariances themselves, propose new mechanisms, or discover that its environmental assumptions are inadequate. In this sense, IRM exemplifies a confirmatory methodology—seeking evidence for existing invariance beliefs—unless embedded in a broader conjecture–criticism–revision loop that supplies operators for apparatus change.

\subsection{Why LLMs Can Appear to Create Knowledge}

Large language models often strike us as if they are creating new explanatory knowledge during conversation. Our claim is that this appearance stems from the special role of the human interlocutor as a general intelligence who actively shapes the interventional distribution over dialogues. In effect, the human supplies conjectures, crafts targeted probes, performs criticism, integrates extra-linguistic information, and selectively preserves successful lines of thought. The resulting transcript distribution is therefore not the model's passive next-token law but an interventionally filtered distribution induced by human actions.

Formally, let conversation alternate between human ($H_t$) and model ($L_t$) turns. A platonic, purely observational view would treat the joint as
\[
p(H_1,L_1,H_2,L_2,\dots)=\prod_t p\big(H_t\mid H_{<t},L_{<t}\big)\;p\big(L_t\mid H_{\le t},L_{<t}\big).
\]
In reality, the human implements interventions $\mathrm{do}(H_t=h_t^\ast)$ chosen by a general-intelligence policy $\pi_H$ that depends on private memory $M_t$ (notes, background knowledge, tools) and on criticism of prior turns:
\begin{align}
 h_{t}^\ast &\sim \pi_H(\,\cdot\mid H_{<t},L_{<t},M_{t-1}\,), \\
 M_t &= \mathrm{Update}\big(M_{t-1},\, H_t, L_t, \mathrm{Critique}(H_t,L_t)\big).
\end{align}
The model, with fixed parameters $\theta$, responds by sampling from
\[
p_\theta\!\left(L_t \;\middle|\; \mathrm{do}(H_{1:t}{=}h_{1:t}^\ast),\,L_{<t}\right).
\]
Two selection effects follow. First, the query effect: humans steer the dialogue into regimes that expose or repair implicit errors (interventions far off the model's training support). Second, the post-selection effect: humans preferentially keep, quote, and build upon successful continuations while discarding failed branches, effectively conditioning the visible transcript on a success predicate $S$. The distribution of published or remembered conversations is thus
\[
p_{\mathrm{visible}}(\text{transcript}) \;\propto\; p_\theta(\text{transcript}\mid \mathrm{do}(H{=}h^\ast_{1:T}))\;\mathbf{1}\!\left[S(\text{transcript},M_T)\right],
\]
which is neither purely observational nor stationary: it is co-authored by an intervening, knowledge-creating agent.

\begin{example}[Human-shaped interventional dialogue]\label{ex:human-llm-intervention}
Let $H_t,L_t$ denote human/model turns. Under human guidance,
\[
L_t \sim p_\theta\!\left(L_t \mid \mathrm{do}(H_{1:t}{=}h_{1:t}^\ast),\,L_{<t}\right),\qquad
h_{t+1}^\ast \sim \pi_H\!\left(\cdot\,\middle|\,H_{\le t},L_{\le t},M_t\right).
\]
Here $\pi_H$ conducts conjecture $\rightarrow$ criticism $\rightarrow$ revision at the dialogue level: it reformulates prompts, injects external facts, and asks counterfactual ``what-if" questions that the model never learns from in the sense of parameter change. Three implications follow:

\begin{enumerate}[label=(\roman*),leftmargin=*]
\item \textbf{Borrowed intelligence.} The system (human$+$LLM) can create explanatory knowledge because the human supplies the conjecture\textendash criticism loop and updates $M_t$. The model, holding $\theta$ fixed, supplies conditional samples and pattern completions; any apparent ``insight" lives in $M_t$, not in $\theta$.

\item \textbf{Interventional filtering.} The sequence of human prompts constitutes a rich family of interventions that push the dialogue distribution far from $P_{\rm obs}$. Failures are turned into further interventions, so implicit errors become explicit and corrigible; successes are amplified by continued exploration along fruitful branches.

\item \textbf{Success-biased transcripts.} Because humans keep and propagate successful branches (and often omit failed ones), public artifacts (notes, posts, papers) exhibit a rising signal of apparent competence over time, even if the model parameters never changed. This creates the appearance of on-the-fly knowledge creation by the model.
\end{enumerate}
\end{example}

\subsection{Training vs Teaching}
\textit{Training}, the dominant hypothesis-closed process by which existing artificial intelligences learn, is suited to transmitting existing explanatory knowledge but not to generating new explanatory knowledge, which requires error correction beyond any single loss function. We describe \textit{teaching} in the same sense it applies to humans: as encouraging an intelligent system to learn through an open-ended, iterative process of conjecture and criticism, with the generation of new explanatory knowledge as the goal.

\section{Three Error-Centric Questions for AGI}\label{sec:error_questions}

The limitations of purely inductive learning point to the need for a deeper epistemological shift. General intelligence requires the capacity to actively create new knowledge, and this implies a capacity for conjecture and criticism. To operationalize this shift, we propose three diagnostic questions that any candidate theory of general intelligence must address. These questions serve as motivation for developing new structural principles (Section \ref{sec:principles}) rather than as problems to be directly solved.

\paragraph{Question 1: The Evolution of Explicit and Implicit Errors}

How does the diversity of possible errors, both explicit and implicit, evolve as the agent executes its sequence of actions or computation steps? More specifically, how does learning affect the diversity of future errors?

Explicit errors manifest in observable failures: a robot falls when sitting, an LLM asserts a falsehood. But these failures often reflect deeper implicit errors, flawed internal representations, heuristics or world models, which become visible only upon execution. Each implicit error may enable many explicit manifestations, varying by context. The agent's own computations determine which errors remain latent and which are made explicit.

Discovering and correcting implicit errors  requires an ability to reason counterfactually when heuristics (e.g., those programmed by evolution or learned via ERM/IRM) cannot capture the specific knowledge required to adequately solve a problem. 

\paragraph{Question 2: Hypothesis Reach and Unreachable Errors}

What hypotheses are available to the agent, and what classes of errors remain inherently unreachable under those hypotheses?

The hypothesis space consists of candidate mappings, programs, or relational structures over the existing representation space that the system can construct, evaluate, and update.  Some errors are unreachable not because of insufficient data, but because the current hypothesis space cannot express, and subsequently improve, certain conjectures. 

The Platonic Representation Hypothesis and ERM assume that given enough data, representations will converge to capture all relevant structure. But this presupposes that the initial hypothesis space is adequate, that all important distinctions are already expressible. If the agent's hypotheses only encode correlational structure, then no amount of observational data will reveal which associations are spurious. Many errors which conflate correlation with causation are unreachable within such a hypothesis space.

\paragraph{Question 3: Conjecture and Criticism Capacity}

How does the agent generate new hypotheses that extend its capacity to detect and correct previously unreachable errors? And how are these hypotheses tested, revised, or discarded?

Conjecture and criticism form the epistemic engine of general intelligence. To function adaptively, an agent must not only revise parameters within a fixed model but invent new structures\textemdash causal, analogical, or otherwise\textemdash that allow for deeper understanding. This requires both a generative mechanism for producing new candidate hypotheses and a critical mechanism for evaluating them against errors. Most learning systems, especially those trained under empirical risk minimization, operate within a fixed hypothesis class. They cannot escape that space without external intervention. But general intelligence requires endogenous hypothesis space revision: a capacity to detect the inadequacy of current models and propose structural alternatives. This process allows the agent to transform unreachable errors into reachable ones, thereby converting failure into epistemic growth. 

\section{Structural Principles Motivated by Error-Centric Questions}\label{sec:principles}

The three error-centric questions motivate the development of structural principles that can guide the design of systems capable of general intelligence. These principles emerge from considering what constraints would enable a system to better exploit its evolving error landscape, extend its hypothesis space reach, and expand its capacity for conjecture and criticism. These structural principles are the foundation of \textit{Causal Mechanics}, our proposal for a mechanisms focused program for causality aware learning that treats hypothesis-space change as a first-class operation and admits probabilistic structure when useful. 

The Locality\textendash Autonomy Principle (LAP) and geometric Independent Causal Mechanisms (ICM) emerge from considering Questions 1\textendash 2: how can we ensure that interventions propagate correctly through causal structure while maintaining the modularity needed to detect and correct errors locally? The Compositional Autonomy Principle (CAP) emerges from considering Question 3: how can analogical reasoning maintain structural integrity while enabling the transfer and composition of knowledge across domains? We also recast Independent Causal Mechanisms (ICM) as a gauge\textendash invariant separability condition with commuting flow witnesses, formulate LAP in differential form via Lie derivatives, and introduce CAP with concrete diagnostics. Part II will operationalize these principles through Energy\textendash Structured Causal Models (E\textendash SCMs).

\subsection{Structural Principles: Locality, Autonomy, and Independent Mechanisms}\label{sec:structural_principles}

This section develops the structural commitments that make causal modeling actionable. We begin with the baseline semantics of structural causal models (SCMs): modularity and invariance under interventions. These commitments are formalized as the Locality\textendash Autonomy Principle (LAP), which captures the idea that each mechanism can be varied independently and that non-descendants remain unaffected by interventions. 

Building on this, we follow \citep{Schoelkopf2012,Peters2017} by introducing Independent Causal Mechanisms (ICM) (also called the Independent Mechanisms Principle/IMP) as an additional conjecture about the organization of nature. 

We recast ICM in differential-geometric terms as a condition of separability, understood up to gauge reparametrizations. 

The purpose of this section is to clearly separate minimal SCM assumptions (LAP) from conjectural principles (ICM) and to connect them to concrete, testable diagnostics such as gradient or Lie penalties, block-diagonal Fisher/metric witnesses, and commuting flows.

\begin{definition}[Structural causal models \textnormal{(standard)}]
Let $G=(V,E)$ be a directed graph. For each $i\in V$, let $X_i\in\mathcal{X}_i$ and $U_i\in\mathcal{U}_i$. Define $\mathrm{PA}(i):=\{\,j\in V : (j,i)\in E\,\}$ and
$\mathcal{X}_{\mathrm{PA}(i)} := \prod_{j\in \mathrm{PA}(i)} \mathcal{X}_j$ 
(empty product $=\{\ast\}$). An SCM consists of structural assignments
\[
  X_i = f_i\!\big(X_{\mathrm{PA}(i)}, U_i\big),
  \quad f_i:\ \mathcal{X}_{\mathrm{PA}(i)} \times \mathcal{U}_i \to \mathcal{X}_i.
\]
This definition is functional: the model specifies a system of equations and exogenous variables $(U_i)$, but not yet a probability law. Endowing $\mathbf{U}=(U_i)_{i\in V}$ with a joint distribution yields a probabilistic SCM. Cycles are permitted, though existence and uniqueness of solutions then require extra conditions.
\end{definition}

\begin{definition}[Markovian and semi-Markovian SCMs \textnormal{(standard, after Pearl)}]
An SCM is \emph{Markovian} if it is acyclic and the $U_i$ are mutually independent, so that each mechanism's exogenous noise is self-contained. It is \emph{semi-Markovian} if acyclicity holds but the $U_i$ may be dependent, in which case dependencies capture latent confounding (often depicted with bidirected edges).
\end{definition}

\begin{definition}[Baseline semantics: modularity and invariance \textnormal{(standard)}]
An SCM encodes two basic commitments:
\begin{enumerate}
  \item \textbf{Modularity/autonomy.} Each $(f_i,U_i)$ is an autonomous module: modifying $X_A$'s mechanism (including by $\mathrm{do}(X_A{=}a)$) leaves all other mechanisms unchanged.
  \item \textbf{Invariance of non-descendants (Markovian case).} If the SCM is Markovian, then for any non-descendant $X_i$ of $X_A$, interventions on $X_A$ do not affect $X_i$'s distribution: $P(X_i\mid \mathrm{do}(X_A{=}a))=P(X_i)$.
\end{enumerate}
Together with acyclicity and independent noises, these commitments entail the usual DAG Markov factorization, but they are logically prior to it.
\end{definition}

\begin{definition}[ICM / IMP \textnormal{(traditional)}]
Beyond SCM semantics, the ICM principle asserts that each child mechanism is independent of the process generating its parents. Standard formalizations use algorithmic or minimum-description-length (MDL) independence between ``cause" and ``mechanism." ICM is additional to SCM semantics and underlies identifiability results and invariance-based tools.
\end{definition}
In this paper, `MDL for causality' refers to a scoring principle used (i) to compare the two bivariate directions $X\!\to\!Y$ vs.\ $Y\!\to\!X$ under an ICM/additive-noise assumption, and (ii) to score DAGs via $L(G)+L(\theta\mid G)+L(\text{data}\mid G,\theta)$ (typically yielding a Markov equivalence class absent further assumptions)—that is, MDL selects among causal hypotheses but does not itself supply interventional semantics.

While ICM is often operationalized via algorithmic independence of cause and mechanism—the Kolmogorov-complexity statement that the shortest joint description of the marginal and the conditional satisfies
\(K(p_X,p_{Y\mid X}) \approx K(p_X)+K(p_{Y\mid X})\) (up to an \(O(1)\) term)—practical work replaces \(K(\cdot)\) by MDL two-part code lengths, comparing \(L(p_X)+L(p_{Y\mid X})\) to \(L(p_Y)+L(p_{X\mid Y})\). In Appendix~\ref{sec:ct-tdl} we sketch a speculative constructor-theoretic alternative: CT-TDL prices mechanisms by the minimal physical resources needed to realize the task to a given accuracy and reliability, rather than by code length, and thus selects the causal direction with lower task cost.

\paragraph{Mechanism notation.}
For node $i$ with parents $\mathrm{PA}(i)$, write the (parametric) mechanism
\[
\mathcal{M}_i:\ \mathcal{X}_{\mathrm{PA}(i)}\times \mathcal{U}_i \times \Theta_i \to \mathcal{X}_i,
\qquad
x_i=\mathcal{M}_i(x_{\mathrm{PA}(i)},u_i;\theta_i),
\]
where $\theta_i\in\Theta_i$ are the parameters of mechanism $i$. Let $\xi_A$ denote the state-flow vector field induced by varying $X_A$ (holding parameters fixed), and let $\Xi_A$ denote the parameter-flow vector field on $\Theta_A$ (varying $\theta_A$ with states fixed).

\begin{definition}[Locality\textendash Autonomy Principle (differential form, \textnormal{new})]
Within an SCM, for any $X_A$ and $X_i$ with $i\notin\mathrm{Desc}(X_A)$:
\begin{align*}
  \mathcal{L}_{\xi_A}\,\mathcal{M}_i &= 0 \quad &\text{(locality: flows of $X_A$ do not affect non-descendants)},\\
  \mathcal{L}_{\Xi_A}\,\mathcal{M}_i &= 0 \quad &\text{(autonomy: perturbations of $\theta_A$ leave other mechanisms unchanged)}.
\end{align*}
Here $\mathcal{L}$ denotes the Lie derivative. For scalar functions this reduces to the directional derivative, and for vector fields to the commutator.
\end{definition}

\begin{definition}[ICM / IMP (geometric, gauge-invariant, \textnormal{new})]
ICM is a structural separability condition, defined up to smooth reparametrizations (gauge) of upstream and child parameters. At node $i$ it requires:
\begin{enumerate}
  \item \textbf{Structural independence.} In some adapted chart (a local coordinate system on parameter space chosen to reflect the structure), holding $\mathrm{PA}(i)$ fixed, $\mathcal{M}_i$ does not vary with upstream parameters: $\partial_{\theta_{\mathrm{PA}(i)}}\,\mathcal{M}_i = 0$.
  \item \textbf{Separability.} $\mathcal{S}_i \subseteq \Theta_{\mathrm{PA}(i)} \times \Theta_i$ admits a local product structure; equivalently, there exist coordinates in which constraints factor as $\big(C_{\theta_{\mathrm{PA}(i)}}(\theta_{\mathrm{PA}(i)}),\, C_{\theta_i}(\theta_i)\big)$. Commuting parameter flows provide a coordinate-free witness.
\end{enumerate}
These conditions are jointly necessary and sufficient for local ICM, understood modulo reparametrizations of $( \theta_{\mathrm{PA}(i)},\theta_i)$. Practical witnesses include block-diagonality of a chosen metric on parameter space (e.g., the Fisher information under a specified observational model), vanishing cross-partials in an adapted chart, or commuting parameter flows (vanishing Lie brackets).
\end{definition}

\begin{remark}[Coordinate-free witness: commuting flows]
As a coordinate-free witness of ICM, we use commuting parameter flows.
Let $\mathcal D_{\mathrm{PA}}$ (upstream/parent) and $\mathcal D_i$ (child) be the smooth distributions
generated by the respective parameter–flow vector fields on
$\Theta_{\mathrm{PA}(i)}$ and $\Theta_i$.
Under mild regularity (smoothness, constant rank on a neighborhood, and complementary spans),
the following are equivalent locally:
(i) there exist product coordinates $(\theta_{\mathrm{PA}(i)},\theta_i)$ in which the child mechanism is
insensitive to upstream parameters, $\partial_{\theta_{\mathrm{PA}(i)}} \mathcal M_i = 0$;
(ii) the flows within $\mathcal D_{\mathrm{PA}}$ and within $\mathcal D_i$ are integrable and the two families
of flows commute, i.e.
\[
[X_{\mathrm{PA}},X'_{\mathrm{PA}}]=0,\quad
[Y_i,Y'_i]=0,\quad
[X_{\mathrm{PA}},Y_i]=0
\]
for all $X_{\mathrm{PA}},X'_{\mathrm{PA}}\in\Gamma(\mathcal D_{\mathrm{PA}})$ and
$Y_i,Y'_i\in\Gamma(\mathcal D_i)$.
($\Gamma(\mathcal D)$ denotes the set of smooth vector-field sections of $\mathcal D$, and equivalence follows from the Frobenius theorem applied to two complementary, commuting foliations.)
\end{remark}

\begin{remark}[On adapted charts and gauge invariance]
An \emph{adapted chart} means a local coordinate system chosen so that the independence condition is manifest. We are not saying that independence requires $\partial_{\theta_{\mathrm{PA}(i)}} \mathcal{M}_i=0$ in every possible coordinate system (which would be too strong and not gauge-invariant). Rather, independence means that there exists some local chart (obtained by a smooth reparametrization) in which this condition holds. If no such chart exists, the coupling is structural — it reflects a real constraint or interaction, not just a bad parametrization.
\end{remark}

\begin{remark}[Structural vs. MDL independence]
ICM is often operationalized by MDL additivity, asking that the code length of parents plus child be (approximately) additive. Our geometric recasting demands structural decomposability: insensitivity to upstream parameters (in an adapted chart) and a feasible set with product structure. This typically entails additive code lengths for universal codes aligned with the factorization, but not vice versa: MDL additivity can hold or fail depending on coding choices even when hidden constraints couple $(\theta_{\mathrm{PA}(i)},\theta_i)$.
\end{remark}

While LAP expresses the minimal structural commitments required by the SCM formalism, ICM is stronger and more conjectural: it posits that mechanisms are not co-adapted but structurally separable. We advance ICM here as a synthetic conjecture rather than an inductive generalization, i.e., a falsifiable structural claim whose failures localize missing mechanisms and prompt revision. Its testable content lies in structural diagnostics: the existence of adapted charts with $\partial_{\theta_{\mathrm{PA}(i)}} \mathcal{M}_i=0$,
approximate block–diagonality of a chosen metric on $\Theta_{\mathrm{PA}(i)}\times\Theta_i$
(e.g., the Fisher information),
 and cross-environment invariances. When these diagnostics fail, they localize missing structure and prompt model revision; when they persist, they demarcate intrinsic-coupling regimes where modular intervention semantics should not be assumed.

It is important to note that in nature, ICM often fails systematically and these failures require causal explanation: in feedback control, coupling occurs because controllers map observed outputs to control actions; in conservation systems, coupling emerges through constraint-enforcement mechanisms; and in co-evolution, coupling results from ecological interactions linking fitness landscapes. These explanations reveal that ICM failures often stem from missing causal structure rather than fundamental non-modularity. The controller, conservation law, or fitness interaction represents an omitted mechanism.

\emph{Meta-conjecture.} Better explanations tend toward ICM compatibility: for many phenomena, augmenting the model with the right latent mechanisms restores (approximate) separability so that, in an adapted chart on the enlarged parameter manifold, cross terms shrink and non-descendant invariances reappear. Nevertheless, some couplings remain intrinsic, e.g., non-integrable constraints, critical phenomena, or quantum entanglement relative to the subsystem partition defined by decoherence, marking real limits to modularization.

\subsection{Analogical Reasoning}\label{sec:analogical_reasoning}

General intelligence requires not only causal conjecture and criticism\textemdash reasoning about how the world evolves through mechanisms\textemdash but also analogical reasoning: recognizing that two situations share the same pattern of relations even when the corresponding objects differ and even when the pattern lives at the level of relations themselves (relation $\leftrightarrow$ relation). People routinely transfer a solution from one domain to another (electrical circuits $\leftrightarrow$ fluid flow; family trees $\leftrightarrow$ corporate hierarchies) by matching how things are related, not the surface features. Analogies as understood here depend either explicitly or implicitly on causal conjectures and may serve as the inspiration for new causal conjectures and for new modes of criticism of existing explanations. This section introduces the Compositional Autonomy Principle (CAP), an independence principle for analogical reasoning that plays a role analogous to the LAP in causality.

\subsubsection{Ingredients and Recipes: The Structure of Analogies}

To make the formal treatment concrete, think of an analogy between two domains $A$ and $B$ (such as family trees and corporate hierarchies) as having several parts. First, each domain has a small set of \textit{primitive} operations or relations---the basic building blocks. We call this set of primitives the \textit{signature} $\Sigma$. For example, in the family domain the signature might contain just the ``parent'' relation, while in the company domain it contains the ``manager'' relation.

Second, we form more complex structures by composing these primitives according to \textit{recipes}. A recipe is a syntactic expression (a term $T$) that specifies how to wire primitives together. For instance, ``apply the parent relation twice in sequence'' is a recipe that yields the grandparent relation. The recipe itself is just syntax---a set of instructions. The \textit{arity} $k(T)$ of a recipe $T$ tells us how many inputs it expects (e.g., the grandparent recipe takes two inputs: a child and a potential grandparent).

Third, once we fix concrete data and parameters in domain $A$, each recipe $T$ determines a \textit{realized map} $[\![ T ]\!]^A$. This is the actual function you get when you follow the recipe's instructions with the specific primitives of domain $A$. It takes $k(T)$ entities from $A$ as input and returns a result (perhaps another entity, a truth value, or a number).

Fourth, an analogy between domains $A$ and $B$ consists of two mappings. The \textit{entity translator} $\Phi\colon A \to B$ converts individual entities from domain $A$ into corresponding entities in domain $B$ (e.g., a person becomes an employee). When a recipe requires multiple inputs, we write $\Phi^{\times k}$ to denote that $\Phi$ acts componentwise on all $k$ inputs. The \textit{symbol correspondence} $F$ maps primitive symbols in $\Sigma_A$ to primitive symbols in $\Sigma_B$ (e.g., $F(\mathrm{parent}) = \mathrm{manager}$). Because $F$ acts on symbols, it extends naturally to recipes: if $T$ is a recipe in domain $A$, then $F(T)$ is the analogous recipe in domain $B$ obtained by replacing each primitive symbol according to $F$.

The core question is whether these mappings preserve the compositional structure. Does computing in domain $A$ and then translating to $B$ give the same result (up to a small error) as translating the inputs first and then computing in domain $B$? In symbols, we ask whether
\[
\Phi\bigl([\![ T ]\!]^A(x)\bigr) \approx [\![ F(T) ]\!]^B\bigl(\Phi^{\times k(T)}(x)\bigr).
\]
When this holds for all relevant recipes $T$ and inputs $x$, the analogy is \textit{compositionally consistent}: map-then-compose matches compose-then-map. The Compositional Autonomy Principle (CAP) formalizes the conditions under which this consistency is maintained during learning.
\subsubsection{Concrete Example}
Let's start with a binary example. Consider the ``family $\leftrightarrow$ company'' analogy. Let the primitive in $A$ be the binary relation $\mathrm{parent}_A(x,y)$ (``$y$ is a parent of $x$''), and in $B$ the binary relation $\mathrm{manager}_B(u,v)$ (``$v$ manages $u$''). A recipe (term) $T$ can be the composition $\mathrm{grandparent}_A := \mathrm{parent}_A \circ \mathrm{parent}_A$; its arity is $k(T)=2$ because it takes a pair $(x,z)$ and returns a truth value (``$z$ is a grandparent of $x$''). The realized map $[\![ T ]\!]^A$ is the function that, given $(x,z)$, checks whether there exists $y$ with $\mathrm{parent}_A(x,y)$ and $\mathrm{parent}_A(y,z)$. The entity translator $\Phi\colon A\to B$ sends each person to an employee (and acts on pairs as $\Phi^{\times 2}(x,z)=(\Phi(x),\Phi(z))$), while the primitive correspondence $F$ maps symbols by $F(\mathrm{parent})=\mathrm{manager}$ and therefore $F(T)=\mathrm{grandmanager}_B := \mathrm{manager}_B \circ \mathrm{manager}_B$. The analogy-consistency check becomes
\[
\Phi\bigl([\![ T ]\!]^A(x,z)\bigr) \approx [\![ F(T) ]\!]^B\bigl(\Phi(x),\Phi(z)\bigr),
\]
which, for predicates, means the disagreement rate between ``$z$ is a grandparent of $x$'' and ``$\Phi(z)$ is a grandmanager of $\Phi(x)$'' is small. 

For a numeric example, let $A$ be shopping calculations in dollars with primitives $\mathrm{multiply}$ (price $\times$ quantity), $\mathrm{discount}$ (apply percentage off), and $\mathrm{sum}$ (total cost), and let $B$ be the same shopping calculations in euros with the same primitives. A term $T$ might be ``buy 3 items at \$15 each, apply a 20\% discount, then add a \$5 shipping fee,'' so $k(T)$ reflects the number of inputs (item price, quantity, discount rate, shipping). The realized map $[\![ T ]\!]^A$ computes the final dollar amount; $\Phi$ converts dollars to euros using the exchange rate, and $F$ preserves the arithmetic operations: $F(\mathrm{multiply})=\mathrm{multiply}$, $F(\mathrm{discount})=\mathrm{discount}$, etc. The CAP equation then tests that performing the shopping calculation in dollars and converting to euros matches converting each input to euros first and computing there, up to a small numeric residual (which might arise from rounding, transaction fees, or exchange rate fluctuations).

\subsubsection{Why CAP?}

Having fixed what an analogy is as a structure-preserving map $\Phi$, the next question is what constraints on learning keep that structure intact as the system changes. The Compositional Autonomy Principle (CAP) addresses exactly this issue by specifying the conditions under which analogical structure is maintained rather than eroded by training.

Three characteristic forms of degradation motivate CAP. First, non-use coupling occurs when updating the parameters of one primitive silently alters composites that never invoke it, introducing spurious cross-talk. Second, law drift arises when optimization improves a task objective at the cost of violating the equations that define the small function algebra of the domain, such as associativity, symmetry, or conservation-like constraints, thereby breaking systematic transfer. Third, a previously valid analogy may degrade so that map-then-compose no longer agrees with compose-then-map; in symbols, $\Phi\circ f_A$ ceases to match $f_B\circ \Phi^{\times k}$ up to a small residual, even though in-domain performance remains unchanged. These are structural errors rather than pointwise prediction mistakes, and they call for structural safeguards.

CAP is an independence principle at the level of symbols and their compositions. It requires locality in the sense that changing the parameters of a primitive affects only those composites that actually use that primitive. It requires law preservation in the sense that the declared equations that endow the signature with its algebraic character remain satisfied and are insensitive to updates to unrelated primitives. It requires analogy consistency in the sense that, for the operations and relations covered by the analogy, the computation obtained by first composing in domain $A$ and then translating matches the computation obtained by first translating and then composing in domain $B$, up to a small and measurable residual. Together these requirements maintain the very structure that $\Phi$ is intended to preserve while learning proceeds.

The formal statement of CAP instantiates these ideas quantitatively: small gradients of non-using composites with respect to a primitive's parameters express locality (for a composite $T$ that does not contain symbol $\sigma$, one enforces $\nabla_{\theta_\sigma} [\![ T ]\!] \approx 0$); small residuals on the target equations express law preservation; and small differences between $\Phi\circ f_A$ and $f_B\circ \Phi^{\times k}$ on held-out terms express analogy consistency. The details mirror the style of identifiability and invariance diagnostics used earlier in the paper.

CAP extends the error-centric view by turning silent degradations of analogical structure into explicit, reachable errors that can be detected, criticized, and corrected. It also reduces representational slack that preserves in-domain loss while scrambling transfer, thereby complementing the discussion of fractured and entangled representations. We next give a compact formal statement of CAP and its associated diagnostics.

\subsubsection{CAP: Core Statement}
The following instantiates the intuition—local edits stay local, algebraic laws remain stable, and map-then-compose agrees with compose-then-map—into quantitative conditions.
\paragraph{Setting.} Let $\Sigma$ be a finite signature of primitive operations/relations (``primitives'') with parameters $\theta=\{\theta_\sigma: \sigma\in\Sigma\}$. A domain $D\in\{A,B\}$ interprets $\Sigma$ as an algebra that maps each syntactic term $T$ (a composition tree over $\Sigma$) to a realized map $[\![ T ]\!]_{\theta}^{D}$. A symbol correspondence $F\colon\Sigma_A\to\Sigma_B$ extends homomorphically to terms $T\mapsto F(T)$, and $\Phi\colon A\to B$ maps entities (componentwise on tuples, written $\Phi^{\times k}$ for $k$ inputs).

\paragraph{Compositional Autonomy Principle (CAP).} CAP asserts three structural requirements:
\begin{enumerate}
  \item \textbf{Locality.} If $T$ contains no occurrence of primitive $\sigma$, then changing $\theta_\sigma$ does not change $[\![ T ]\!]^D$.
  
  \textit{Operational witness:} small non-use Jacobians $\nabla_{\theta_\sigma} [\![ T ]\!]^D \approx 0$ whenever $\sigma\notin T$.
  \item \textbf{Law stability.} A designated set of identities (``laws'') over terms remains approximately satisfied in $D$ and is insensitive to parameters of primitives that do not appear in those identities.
  \item \textbf{Analogy consistency.} For covered terms $T$, composing in $A$ and then mapping agrees with mapping and then composing in $B$ (up to a small residual):
  \[
  \Phi\bigl([\![ T ]\!]_{\theta}^{A}(x)\bigr) \approx [\![ F(T) ]\!]_{\theta}^{B}\bigl(\Phi^{\times k(T)}(x)\bigr).
  \]
\end{enumerate}
\noindent Intuitively: locality prevents spurious cross-talk; law stability preserves the domain's algebraic character (associativity, symmetries, conservation-like constraints); analogy consistency encodes the essence of ``map-then-compose $\approx$ compose-then-map.''

\subsubsection{Witnesses and Diagnostics for CAP}
\paragraph{Primitives, terms, and realized maps.} Each primitive is a parametric module $R_\sigma(\cdot;\theta_\sigma)$ on entity vectors in space $V$. A syntactic tree $T$ specifies how primitives are wired; evaluating $T$ under parameters $\theta$ yields a realized map $[\![ T ]\!]_{\theta}^{D}\colon V^{\otimes k(T)}\to V^{\otimes m(T)}$. This is the same content previously introduced as $\mathrm{Eval}_\theta(T)$.

\paragraph{Locality diagnostic.} For any primitive $\sigma$ and any composite $T$ with $\sigma\notin T$,
\[
\mathcal{L}_{\text{loc}}(\sigma,T) := \mathbb{E}_{X}\,\big\|\nabla_{\theta_\sigma}\, [\![ T ]\!]^{D}_{\theta}(X)\big\|^{2} \le \varepsilon_{\text{loc}}.
\]
Averaging over inputs and such $T$ yields a scalar locality score per primitive.

\paragraph{Law stability diagnostic.} Let $\{\Phi_{\ell}(T_1,\dots,T_m)=0\}$ be target identities. Define a residual at $\theta$ by
\[
\mathcal{E}_{\ell}(\theta) := \mathbb{E}_X\,\big\|\mathrm{Eval}_{\theta}\big(\Phi_{\ell}\big)(X)\big\|,\quad \sum_{\ell}\mathcal{E}_{\ell}(\theta)^2 \le \varepsilon_{\text{law}}.
\]
Insensitivity to unrelated primitives is expressed as $\|\partial\mathcal{E}_{\ell}/\partial\theta_\sigma\|\le \varepsilon_{\text{ins}}$ whenever $\sigma$ does not occur in $\Phi_{\ell}$.

\paragraph{Analogy consistency diagnostic.} For a covered primitive $f$ of arity $k$, or any composite $T$ of arity $k(T)$, define
\[
H_f(\Phi) := \mathbb{E}_{x\in A^{k}}\, d\!\bigl(\,\Phi\big(f_A(x)\big),\, f_B\big(\Phi^{\times k}(x)\big)\,\bigr),
\]
\[
H_T(\Phi) := \mathbb{E}_{x}\, d\!\bigl(\,\Phi\big([\![ T ]\!]^{A}_{\theta}(x)\big),\, [\![ F(T) ]\!]^{B}_{\theta}\big(\Phi^{\times k(T)}(x)\big)\,\bigr),
\]
with a norm or disagreement rate $d(\cdot,\cdot)$. Small values indicate homomorphism-like behavior.

\subsubsection{Failure Modes and What Catches Them}

\begin{enumerate}
  \item \textbf{Spurious analogy.} Surface alignment but law violations: high $\sum_{\ell}\mathcal{E}_{\ell}^2$.
  \item \textbf{Training drift (analogy erosion).} Homomorphism residual $H_T(\Phi)$ grows over time despite stable in-domain loss.
  \item \textbf{Non-use coupling.} Edits to $\theta_\sigma$ perturb composites not using $\sigma$: elevated non-use Jacobians.
\end{enumerate}

\subsubsection{Relation to LAP and ICM}

ICM concerns separability \emph{within} a domain (structural independence of child mechanisms from upstream parameterization, up to gauge). CAP concerns separability \emph{under} cross-domain translation: does a structure-preserving correspondence exist and stay stable during learning? ICM can hold while CAP fails (mechanisms are separable in $A$ but no faithful $\Phi,F$ make $A\to B$ compositional), or CAP can approximately hold despite local ICM violations if a higher-level algebra remains stable. This complements Section~\ref{sec:structural_principles}: LAP gives modular interventions; ICM posits separability; CAP preserves compositional structure needed for transfer.

\subsubsection{When CAP is Informative}

CAP residuals constrain models only when (i) the term set covers the constructions of interest (not just primitives but key composites), (ii) inputs cover a diversity of entities and contexts, and (iii) $F$ and $\Phi$ are not over-permissive. Multi-sorted settings apply CAP sortwise; $\Phi^{\times k}$ acts componentwise on typed tuples. Approximate symmetries or conservation laws can be expressed as identities with soft residuals.

\medskip
\noindent\textit{Summary.} CAP asserts that analogical structure is preserved by construction: non-using parts stay inert (locality), the domain's algebra remains stable (law stability), and cross-domain composition commutes with translation (analogy consistency). The witnesses turn these structural claims into measurable residuals that can be used for diagnosis or gentle regularization without collapsing the distinction between a principle and a loss function.

\section{Conclusion}\label{sec:conclusion}

We have argued that progress toward artificial general intelligence is theory limited rather than data or scale limited. Building on the Deutsch–Popper view, we shifted the central obstacles from “more data and compute” to three error centric questions: (i) how explicit and implicit errors evolve under an agent’s actions; (ii) which errors are unreachable within the current hypothesis space; and (iii) how conjecture and criticism expand that space.

Our analysis of the Platonic Representation Hypothesis makes the core point precise: observational adequacy does not secure interventional competence. Further, the Popper–Miller result clarifies why probability raising alone supplies no new explanatory content. In short: Bayes reweights, $\mathrm{do}(\cdot)$ rewires, conjecture proposes hypothesis space-changing moves, and criticism both tests and directs those moves.

Motivated by these findings, we stated structural commitments—not solutions—that make error discovery and correction more tractable: the Locality–Autonomy Principle (LAP) for modular interventions, a gauge invariant formulation of Independent Causal Mechanisms (ICM) for separability, and the Compositional Autonomy Principle (CAP) for preserving analogical structure during learning. These principles support a program in which hypothesis space change is first class: altering intervention semantics where appropriate, refactoring invariances across environments, and introducing new variables and mechanisms when demanded by critical feedback. Criticism here is not only a filter on claims; it localizes violations, prioritizes corrections, and shapes the search over alternative structures.

Finally, the LAP helps to explain catastrophic forgetting: under exact LAP, first-order interference vanishes and under approximate LAP, cumulative drift is bounded by measured locality/autonomy violations (see  Appendix~\ref{sec:appendix-FER-LAP}).

Part~II develops Energy Structured Causal Models (E–SCMs) that operationalize aspects of this program. E–SCMs replace function computation with constraint mechanisms to support probability optional abduction–intervention–prediction, unify static and dynamic settings, allow for cyclic causal graphs, and handle latent space interventions. We provide diagnostics and penalties for violations of LAP, ICM, and CAP.

This work has limits: we do not claim to solve open-ended intelligence. The aim is a coherent scaffold: definitions, principles, diagnostics, and a modeling calculus through which conjecture and criticism can change the hypothesis space and convert unreachable errors into reachable ones.

\newpage
\bibliographystyle{plainnat}
\bibliography{refs}

\newpage
\appendix
\section{Operationalizing Intelligence}
\label{sec:appendix-IntelOp}
The preceding definition identifies intelligence as the efficiency with which a system creates explanatory knowledge. This section outlines three non-exhaustive formalizations of that efficiency. Each functional isolates a distinct aspect of explanatory creation while avoiding conflation with competence. The aim is not to propose a single universal metric but to illustrate how explanatory creation can be made operational in principle. All three measures quantify explanatory gain per unit resource cost rather than performance within a fixed hypothesis space.

Let $\mathcal{R}$ denote a chosen resource basis (time, compute, samples) and $\mathrm{Cost}_{\mathcal{R}}$ its associated cost functional. Let $\mathfrak{K}_0$ and $\mathfrak{K}_1$ represent a system's explanatory knowledge before and after a learning episode. The set $\Delta \mathfrak{K} = \mathfrak{K}_1 \setminus \mathrm{Cn}(\mathfrak{K}_0)$ contains the newly created, non-entailed explanatory commitments, each paired with a test battery $\mathcal{T}(c)$ and severity weights $s(\tau)\in[0,1]$.

\subsection{Explanatory Creation Rate (ECR)}

The explanatory creation rate measures how efficiently new explanatory knowledge is produced and survives criticism. It is defined as
\begin{equation}
\mathrm{ECR}=\frac{1}{\mathrm{Cost}_{\mathcal{R}}}\sum_{c \in \Delta\mathfrak{K}} \Big[ \Big( \sum_{\tau \in \mathcal{T}(c)} s(\tau)\,\mathbf{1}\{c \text{ survives } \tau\} \Big)\,\mathbf{1}\{c \text{ changes a } \mathrm{do}\text{-law}\} \Big]
\end{equation}
It credits only those conjectures that alter interventional structure and withstand severe tests. A research laboratory discovering new causal mechanisms in viral evolution exemplifies a high ECR: a few durable explanations emerging from many conjectures, normalized by experimental cost. A predictive model that achieves accuracy without introducing new mechanisms has ECR = 0, even if the agent’s process for constructing that model scores positively.

\subsection{Counterfactual Reach Expansion (CRX)}

The counterfactual reach expansion quantifies growth in the range of counterfactual questions a system can now answer. Let $\mathcal{Q}$ denote a fixed family of interventional queries and $\mathcal{A}(\mathfrak{K}) \subseteq \mathcal{Q}$ the subset answerable given a knowledge set $\mathfrak{K}$. Then
\begin{equation}
\mathrm{CRX}=\frac{1}{\mathrm{Cost}_{\mathcal{R}}}\sum_{q \in \mathcal{A}(\mathfrak{K}_1)\setminus \mathcal{A}(\mathfrak{K}_0)} w(q)\,\mathbf{1}\{q \text{ validated on new interventions}\}.
\end{equation}
CRX rises when a system's model becomes capable of formulating and evaluating new intervention queries that were previously undefined. A climate model that, after incorporating cloud feedback mechanisms, can now simulate doubling-CO$_2$ scenarios exemplifies a gain in counterfactual reach. The same logic applies to a learner who, after mastering Newtonian mechanics, can now reason about hypothetical worlds beyond direct experience.

\subsection{Structural Edit Yield (SEY)}

The structural edit yield measures the productivity of structural changes to a system's explanatory model. Let $\mathcal{E}$ denote the set of mechanism-level edits proposed, each inducing a change in the model's intervention semantics. For each edit $e$, let $\mathrm{Fail}(e)$ count falsifying tests passed and $\mathrm{Hold}(e)$ indicate survival after a fixed critical horizon. Then
\begin{equation}
\mathrm{SEY}=\frac{1}{\mathrm{Cost}_{\mathcal{R}}}\sum_{e \in \mathcal{E}} \big[\alpha\,\mathrm{Fail}(e)+\beta\,\mathbf{1}\{\mathrm{Hold}(e)\}\big],
\end{equation}
with fixed positive constants $\alpha$ and $\beta$. SEY captures the efficiency with which structural revisions produce enduring explanatory improvement. A biologist who revises a causal diagram of cell signaling to include a feedback loop that resolves prior anomalies exemplifies a high SEY. So does an engineer who introduces an equilibrium constraint into a learning architecture, yielding more interpretable and causally coherent behavior.

\subsection{Interpretation}

ECR, CRX, and SEY address complementary facets of explanatory intelligence. ECR quantifies the rate of explanatory innovation per resource; CRX measures the expansion of reachable counterfactual space; SEY tracks the yield of structural edits that survive criticism. None rely on performance metrics or reward signals internal to a fixed environment. They evaluate the generative and corrective activity that enlarges a system's explanatory domain. While idealized, these functionals illustrate how the creation of explanatory knowledge can be treated as an observable, measurable process rather than a philosophical abstraction.

\section{Catastrophic Forgetting: Relationship Between Gradient Interference, FER, and LAP}
\label{sec:appendix-FER-LAP}

Catastrophic forgetting can be described at two complementary levels: a dynamical level, concerned with the trajectory of parameter updates under gradient descent, and a structural level, concerned with the decomposition of the model into local autonomous mechanisms. The dynamical level explains \emph{how} forgetting arises step by step; the structural level explains \emph{when} it can arise at all.

\paragraph{Gradient dynamics.}
Let $\theta$ denote the model parameters and $R_A(\theta), R_B(\theta)$ the risk functions for two tasks $A$ and $B$. Training on $B$ with a stochastic gradient step $\widehat g_B$ updates
\[
\theta^{+} = \theta - \eta\,\widehat g_B, \qquad \mathbb{E}[\widehat g_B]=g_B=\nabla R_B(\theta).
\]
To first order in $\eta$, the change in $A$’s loss is
% \[
% R_A(\theta^{+})-R_A(\theta) \approx -\eta\,\langle g_A, \widehat g_B\rangle.
% \]
\begin{equation}
\Delta R_A^{(1)} \;:=\; R_A(\theta^{+})-R_A(\theta)
\;\approx\; -\eta\,\langle g_A, \widehat g_B\rangle.
\label{eq:single-step-delta-RA}
\end{equation}
The inner product $\langle g_A, g_B\rangle$ therefore determines whether the update for $B$ helps or harms $A$. When the gradients are aligned, the same descent direction reduces both losses. When they point in opposing directions, the update that benefits $B$ increases $A$’s risk. Averaging over many steps gives a cumulative change proportional to the time-average of this inner product. Persistent negative alignment produces linear growth of the loss on $A$ over training on $B$.

This inner-product account matches existing explanations that view forgetting as gradient conflict or subspace overlap; methods such as PCGrad, OGD, GEM, and related interference-minimization approaches explicitly aim to reduce $\langle g_A, g_B\rangle$ by projection or inequality constraints \citep{yu2020pcgrad,Farajtabar2020OGD,LopezPaz2017GEM,riemer2019mer}. In importance-based methods, the Fisher or related curvature plays a similar role by penalizing movement along sensitive directions, thereby shrinking typical cross-task alignment \citep{Kirkpatrick2017EWC,Zenke2017SI,Aljundi2018MAS}. The geometric quantity $\rho_{A,B}=\cos(g_A,g_B)$ measures this alignment. Catastrophic forgetting occurs when $\rho_{A,B}<0$ on average over the trajectory. The gradient picture is therefore a local dynamical account of interference between tasks.

\paragraph{Structural interpretation.}
The locality–autonomy principle (LAP) describes conditions under which mechanisms within a model do not interfere through their parameters. Each mechanism $\mathcal{M}_i$ occupies a parameter block $\theta_i$ and transforms a subset of variables in the causal graph. Locality states that the output of a composite that does not use $\mathcal{M}_i$ is insensitive to $\theta_i$, implying $\nabla_{\theta_i}[\![T]\!] \approx 0$. Autonomy states that one mechanism’s parameters do not change the behavior of another, implying $\nabla_{\theta_i}\mathcal{M}_j \approx 0$ for $j\neq i$ unless $j$ depends on $i$. Together these imply that the Gauss–Newton or Fisher information matrix is approximately block diagonal. 

Under LAP, gradients for different mechanisms are confined to distinct parameter blocks:
\[
g_A = (g_A^{(1)},\ldots,g_A^{(m)}), \qquad g_B = (g_B^{(1)},\ldots,g_B^{(m)}),
\]
where $g_A^{(i)}$ is nonzero only if task $A$ uses mechanism $\mathcal{M}_i$. The cross-task alignment decomposes as
\[
\langle g_A, g_B\rangle = \sum_i \langle g_A^{(i)}, g_B^{(i)}\rangle.
\]
If $A$ and $B$ depend on disjoint mechanism sets, or if non-use Jacobians vanish, all terms in this sum are negligible and the first-order interference term disappears. In this sense, LAP identifies the structural conditions under which gradient interference cannot occur. This structural perspective is not present in the projection and importance literature, which treats overlap as an optimization issue \citep{yu2020pcgrad,Farajtabar2020OGD,LopezPaz2017GEM,Kirkpatrick2017EWC}. Here forgetting is the quantitative symptom of LAP violation: when mechanisms are not fully local or autonomous, non-use Jacobians leak sensitivity into unintended blocks, creating spurious gradient overlap and positive inner products.

\paragraph{Connection to fractured and unified representations.}
A fractured representation violates autonomy by realizing a single capability across many disconnected parameter regions. The gradients for that capability become widely distributed, so other tasks are statistically more likely to overlap with at least one of those regions. Entanglement, in turn, violates locality by mixing distinct capabilities within the same parameter directions. The idea that overlapping distributed codes drive interference is classical \citep{french1999catastrophic} and is echoed in modern accounts of superposition and polysemantic features that share directions \citep{elhage2022toymodels}, as well as in surveys of continual learning and representational drift \citep{hadsell2020tics}. What is added here is the fracture component and its consequence for the \emph{frequency} of overlaps, and the identification of the UFR as the case where LAP holds approximately: mechanisms are localized, parameters are autonomous, and the Fisher matrix is close to block diagonal. Related modular or structure-growing approaches pursue a similar end operationally but do not supply a causal-structural criterion \citep{li2019learntogrow}.

\paragraph{Complementary views.}
The gradient account is a dynamical explanation of forgetting; LAP provides the structural invariants that make those dynamics either possible or impossible. When LAP is satisfied, interference is geometrically constrained and first-order forgetting vanishes. When it is violated, the non-use and cross-block Jacobians open channels through which gradients for one task can pull parameters in directions that increase another’s loss. Thus the two perspectives are consistent: the gradient picture describes the local forces of interference, and LAP specifies the causal architecture that governs where those forces can act. The alignment with prior accounts lies in the inner-product mechanism and curvature sensitivities \citep{yu2020pcgrad,Farajtabar2020OGD,LopezPaz2017GEM,Kirkpatrick2017EWC}, while the novelty lies in using LAP to derive block-sparse sensitivities and in distinguishing fracture from entanglement as separate sources of persistent interference.

\begin{lemma}[LAP controls cross-task gradient alignment]
\label{lem:lap_alignment_bound}
Partition parameters into mechanism blocks $\theta=(\theta_1,\dots,\theta_m)$ and write the task gradients as $g_A=(g_A^{(1)},\dots,g_A^{(m)})$, $g_B=(g_B^{(1)},\dots,g_B^{(m)})$. Let $S_A,S_B\subset\{1,\dots,m\}$ be the sets of blocks used by the computation graphs for tasks $A,B$. Assume approximate LAP holds with constants $\varepsilon_{\mathrm{loc}},\varepsilon_{\mathrm{aut}}\ge 0$ in the following sense: for any composite $T$ that does not call mechanism $i$, $\|\nabla_{\theta_i}[\![T]\!]\|\le \varepsilon_{\mathrm{loc}}$, and for any pair $(i,j)$ with $j\notin \mathrm{Desc}(i)$, $\|\nabla_{\theta_i}\mathcal{M}_j\|\le \varepsilon_{\mathrm{aut}}$. Then there exists a model-dependent constant $C\ge 1$ (depending on operator norms of local Jacobians) such that
\[
\big|\langle g_A,\,g_B\rangle\big|
\;\le\;
C\sum_{i\in S_A\cap S_B}\|g_A^{(i)}\|\,\|g_B^{(i)}\|
\;+\;
C\big(\varepsilon_{\mathrm{loc}}+\varepsilon_{\mathrm{aut}}\big)\,\|g_A\|\,\|g_B\|.
\]
In particular, if $S_A\cap S_B=\varnothing$ and $\varepsilon_{\mathrm{loc}}=\varepsilon_{\mathrm{aut}}=0$, then $\langle g_A, g_B\rangle=0$.
\end{lemma}

\noindent\textit{Proof sketch.}
Decompose the inner product by blocks: $\langle g_A,g_B\rangle=\sum_i \langle g_A^{(i)},g_B^{(i)}\rangle$. For $i\notin S_A\cap S_B$, the chain rule expresses $g_A^{(i)}$ or $g_B^{(i)}$ as a product of Jacobians along paths that either do not use mechanism $i$ (controlled by $\varepsilon_{\mathrm{loc}}$) or traverse non-descendant links (controlled by $\varepsilon_{\mathrm{aut}}$). Bounding the corresponding operator norms and applying Cauchy–Schwarz yields $|\langle g_A^{(i)},g_B^{(i)}\rangle|\le C(\varepsilon_{\mathrm{loc}}+\varepsilon_{\mathrm{aut}})\|g_A\|\|g_B\|$. Summing over $i$ gives the stated inequality, with $C$ absorbing uniform Lipschitz constants of local Jacobians. The zero-alignment case follows by taking $S_A\cap S_B=\varnothing$ and exact LAP.

\medskip
\begin{corollary}[Exact LAP $\Rightarrow$ no first-order forgetting]
\label{cor:exact-lap-no-first-order}
Under exact LAP with disjoint mechanism support for $A$ and $B$ (i.e., $S_A\cap S_B=\varnothing$ and $\varepsilon_{\mathrm{loc}}=\varepsilon_{\mathrm{aut}}=0$),
the first-order change in $A$’s risk during an update for $B$ vanishes:
\[
\Delta R_A^{(1)} \;\approx\; -\eta\,\langle g_A, g_B\rangle \;=\; 0,
\]
using \eqref{eq:single-step-delta-RA} and Lemma~\ref{lem:lap_alignment_bound}.
\end{corollary}

\medskip
\begin{corollary}[Single-step bound under approximate LAP]
\label{cor:approx-lap-single-step}
Under the conditions of Lemma~\ref{lem:lap_alignment_bound},
\[
\big|\Delta R_A^{(1)}\big|
\;\lesssim\; \eta\,\big(\varepsilon_{\mathrm{loc}}+\varepsilon_{\mathrm{aut}}\big)\,\|g_A\|\,\|g_B\|,
\]
where $\lesssim$ absorbs model-dependent operator-norm constants of local Jacobians. This follows by combining \eqref{eq:single-step-delta-RA} with Lemma~\ref{lem:lap_alignment_bound}.
\end{corollary}

\medskip
\begin{corollary}[Multi-step bound]
\label{cor:approx-lap-multistep}
Let $\theta_{t+1}=\theta_t-\eta\,\widehat g_B(\theta_t)$ be $T$ steps of stochastic gradient descent on $B$ with $\mathbb{E}[\widehat g_B|\theta_t]=g_B(\theta_t)$ and let $R_A$ be $L$-smooth. Then
\[
\mathbb{E}\big[R_A(\theta_T)-R_A(\theta_0)\big]
\;\le\;
-\eta \sum_{t=0}^{T-1}\mathbb{E}\,\langle g_A(\theta_t),g_B(\theta_t)\rangle
\;+\;\tfrac{L}{2}\eta^2\sum_{t=0}^{T-1}\mathbb{E}\|\widehat g_B(\theta_t)\|^2,
\]
and Lemma~\ref{lem:lap_alignment_bound} bounds each $\langle g_A(\theta_t),g_B(\theta_t)\rangle$ by the overlap term on $S_A\cap S_B$ plus a residual proportional to $\varepsilon_{\mathrm{loc}}+\varepsilon_{\mathrm{aut}}$. Thus exact LAP with disjoint mechanisms eliminates first-order forgetting across many steps; approximate LAP makes cumulative forgetting scale with measured LAP violations.
\end{corollary}

\section{Formal Details for the Compositional Autonomy Principle (CAP)}

This appendix adds technical details that clarify how CAP is instantiated in learning settings and how its diagnostics are evaluated, without changing the concepts presented in Section~\ref{sec:analogical_reasoning}.

\subsection{Multi-sorted semantics}

We work with a multisorted signature $\Sigma$ that may contain both functions and relations. Each domain $D\in\{A,B\}$ assigns to each sort $s$ a carrier set $\mathcal X_s^D$, and interprets each primitive symbol $\sigma\in\Sigma$ as a parametric map $R_\sigma(\cdot;\theta_\sigma)$ of the appropriate arity and sorts. A syntactic term $T$ is a composition tree over $\Sigma$. Its realized map $[\![T]\!]_\theta^D$ is obtained by wiring the primitives according to the structure of $T$ and evaluating with parameters $\theta=\{\theta_\sigma\}$. Predicates return values in $\{0,1\}$ or $[0,1]$ on their designated sorts; functions return elements of their output sort. A correspondence $F:\Sigma_A\to\Sigma_B$ preserves arities and sorts and extends to terms by replacing each primitive symbol inside $T$ and keeping the composition pattern unchanged. The entity map $\Phi:A\to B$ acts sortwise; on a $k$-tuple it applies componentwise, written $\Phi^{\times k}(x_1,\dots,x_k)=(\Phi(x_1),\dots,\Phi(x_k))$.

\subsection{Locality as inertial independence}

Locality asks that a primitive that does not appear inside a composite term behaves as if it were inert with respect to that term. Concretely, if a term $T$ contains no occurrence of $\sigma$, then changing $\theta_\sigma$ should not change the output of $[\![T]\!]_\theta^D$. This is measured by a non-use Jacobian
\[
\mathbb{E}_X\,\big\|\nabla_{\theta_\sigma}[\![T]\!]_\theta^D(X)\big\|^2 \approx 0
\quad\text{whenever }\sigma\notin T.
\]
The expectation $\mathbb{E}_X$ is taken over an evaluation distribution on inputs for the carriers of $T$ (held-out or synthetic, fixed for diagnostics). The Jacobian norm can be any fixed operator norm or squared Frobenius norm; conclusions are invariant up to constant factors. Small values certify inertial independence: parameters of a primitive not present in $T$ behave as if at rest with respect to $[\![T]\!]$, so edits do not propagate into unrelated composites.

\subsection{Anti-degeneracy conditions}

To exclude trivial solutions, the translator $\Phi$ is assumed to be injective on the region of interest and bi-Lipschitz on compact subsets. Bi-Lipschitz means there exist constants $0<c\le C<\infty$ such that for all $x,y$ in the region,
\[
c\,\|x-y\|\le \|\Phi(x)-\Phi(y)\|\le C\,\|x-y\|.
\]
This prevents collapse and uncontrolled expansion while leaving geometry otherwise flexible. The correspondence $F$ acts symbolically and preserves arities and sorts; it does not learn per-term shortcuts. These regularity constraints ensure that small analogy residuals are achieved by genuine structural alignment rather than collapse.

\subsection{Metric witness and adapted coordinates}

A chosen metric on the parameter manifold $\Theta=\Phi\times\Theta_i$ provides a witness of separability. A common choice is the Fisher information under a specified observational model. In coordinates adapted to the parent–child split, approximate block-diagonality means that off-block entries between parent and child coordinates are small. Operationally, this indicates that second-order sensitivities couple weakly across the split and aligns with the locality goal. When such coordinates exist locally, the apparent dependence can be removed by reparametrization; when no such chart exists, the coupling is structural rather than a parametrization artifact.

\subsection{Stochastic extension and pushforward distances}

When primitives are stochastic, the realized map $[\![T]\!]_\theta^D$ defines a distribution on the output sort rather than a single value. In this setting analogy consistency compares distributions. For a measurable map $\Psi$ and measure $\mu$, the pushforward $\Psi_\#\mu$ is defined by $\Psi_\#\mu(S)=\mu(\Psi^{-1}(S))$; equivalently, if $X\sim\mu$ then $\Psi(X)\sim \Psi_\#\mu$. Residuals are measured using an integral probability metric $d$, which quantifies how far two probability distributions are from one another under a chosen class of test functions. Common choices include the Wasserstein distance and the Maximum Mean Discrepancy. With this notation, analogy residuals are expressed as
\[
\mathbb{E}\, d\!\left(\Phi_\#\mu_T^A,\; \nu_{F(T)}^B{}_\#\Phi^{\times k(T)}\right)\ \text{ small,}
\]
where $\mu_T^A$ and $\nu_{F(T)}^B$ are the output laws of $[\![T]\!]_\theta^A$ and $[\![F(T)]\!]_\theta^B$.

\subsection{Coverage and identifiability of composites}

CAP residuals are informative on the closure of the term set that actually appears during training and diagnosis. A grammar specifies which composites can be formed from primitives by arity-respecting composition, and the depth of a term is the height of its composition tree. The covered family $\mathcal T$ consists of all terms generable by the grammar up to a maximum depth used during training and diagnostics. If a primitive or a particular composition never occurs within $\mathcal T$, its locality and analogy behavior cannot be tested directly. Practical use therefore requires that $\mathcal T$ generate the composites of interest to the application, so that the measured residuals constrain unseen but related terms built from the same primitives.

\subsection{Generalization under CAP residual bounds}

Assume each primitive $R_\sigma$ is $L_\sigma$-Lipschitz in its inputs and parameters. Suppose for all generating terms $T\in\mathcal T$ and all law constraints $\Phi_\ell$ we have bounds
\[
\mathbb{E}\big\|\nabla_{\theta_\sigma}[\![T]\!]_\theta^D(X)\big\|^2\le\varepsilon_{\mathrm{loc}},\qquad
\mathbb{E}\big\|\mathrm{Eval}_\theta(\Phi_\ell)(X)\big\|\le\varepsilon_{\mathrm{law}},\qquad
\mathbb{E}\, d\big(\Phi([\![T]\!]_\theta^A(X)),\,[\![F(T)]\!]_\theta^B(\Phi^{\times k(T)}(X))\big)\le\varepsilon_{\mathrm{ana}}.
\]
Then for any composite $T'$ of depth $d$ formed from $\mathcal T$ by the same grammar,
\[
\mathbb{E}\, d\!\left(\Phi([\![T']\!]_\theta^A(X)),\,[\![F(T')]\!]_\theta^B(\Phi^{\times k(T')}(X))\right)
\le C(d,\{L_\sigma\})\big(\varepsilon_{\mathrm{loc}}+\varepsilon_{\mathrm{law}}+\varepsilon_{\mathrm{ana}}).
\]
The constant $C(d,\{L_\sigma\})$ depends on depth and the Lipschitz constants, and can be taken to grow at most multiplicatively in $\max_\sigma L_\sigma$ with depth. The bound states that approximate satisfaction of locality, law stability, and analogy consistency on a generating set propagates to unseen composites of bounded depth, with error controlled by the same residuals.

\subsection{Law stability with an explicit evaluator}

Law stability quantifies the preservation of the algebraic equations that define the domain’s small function algebra. For an identity $\Phi_\ell(T_1,\dots,T_m)=0$ such as associativity $T_1\circ(T_2\circ T_3)-(T_1\circ T_2)\circ T_3$, the residual $\mathrm{Eval}_\theta(\Phi_\ell)$ is computed by evaluating each $T_j$ under parameters $\theta$ and taking the resulting difference in the target carrier. For numeric outputs one uses a vector norm; for predicates one uses a disagreement rate.

\subsection{Interpretation of the diagnostics}

Locality quantifies inertial independence: unused primitives do not influence composites that do not invoke them. Law stability quantifies the preservation of the equations that endow the signature with its algebraic character, so that training does not improve task loss by breaking the defining laws. Analogy consistency quantifies whether map–then–compose agrees with compose–then–map across domains, which is the operational meaning of structural analogy in this setting. Together these measurements provide a structural audit: small residuals predict reliable analogical transfer within the span of covered composites, while large residuals identify specific mechanisms, correspondences, or laws that require revision.

\section{Constructor-Theoretic Task Description Length (CT-TDL)}
\label{sec:ct-tdl}

Minimum description length (MDL) provides a pragmatic bridge between data compression and model selection, but it remains tied to a representational framework that presupposes a code, a prior, and a stochastic environment. 

Constructor theory invites a more general formulation in which informational parsimony is expressed as the minimal physical resources required to realize a task to a given accuracy and reliability under the laws that govern a substrate, recovering Kolmogorov complexity when only program length is counted and physical costs are idealized away, and recovering MDL when the substrate is a statistical coding setup with expected codelength as the operative resource.

Let a task $\mathcal T: X \Rightarrow Y$ denote a physically permitted transformation between input and output attributes, and let $\mathcal C$ be the class of constructors available for its realization. Each constructor $\Pi \in \mathcal C$ carries a resource cost measured in a chosen vector of resources $\mathcal R$ (program bits, time, energy, memory, or communication qubits). For a fixed accuracy threshold $\varepsilon$, the constructor-theoretic task description length is defined as
\begin{equation}
\mathrm{CT\text{-}TDL}_{L,S;\,\mathcal C,\mathcal R}(\mathcal T;\varepsilon,r)
= \min_{\Pi \in \mathcal C} \; \mathrm{Cost}_{\mathcal R}(\Pi;\varepsilon,r)
\quad \text{s.t.} \quad \Pi \text{ performs } \mathcal T \text{ within error } \varepsilon \text{ and reliability } r.
\end{equation}
A finite CT-TDL indicates that the task is physically possible with finite resources, while an infinite CT-TDL signifies an impossible transformation (for instance, the cloning of an unknown quantum state). The metric inherits the compositional properties of tasks: for serial compositions, $\mathrm{CT\text{-}TDL}(\mathcal T_2 \circ \mathcal T_1)$ is bounded by the sum of individual costs, and for reusable subconstructors, amortization across tasks reflects the economy of shared mechanisms.

Traditional measures arise as regime-specific limits of CT-TDL: Kolmogorov complexity and MDL as described; Shannon (and von Neumann) entropy when the task is asymptotic lossless source coding for classical (and quantum) sources and the counted resource is the coding rate (expected bits or qubits per symbol). In this view, informational simplicity is not defined relative to a symbolic encoding alone but to the physics of construction: a task is simple when it can be achieved with minimal physical effort under the governing laws. 

In causal modeling, the same concept may yield a natural criterion for directionality. Given competing tasks $\mathcal T_{X\to Y}$ and $\mathcal T_{Y\to X}$ that represent alternative mechanisms across environments, the causal direction is the one with the lower robust CT-TDL, defined as the supremum of task costs over allowed interventions. Causal mechanisms are thereby characterized by their reusability and stability under change: they are the transformations that remain physically cheap to reconstruct when environments vary. This constructor-theoretic generalization preserves the spirit of minimum description length but grounds it in the modal structure of physical law rather than the syntax of a code.
\paragraph{Related work and novelty.}
There is an extensive line of work linking description length and causality through the algorithmic viewpoint. The independence of cause and mechanism and the algorithmic Markov condition \citep{JanzingSchoelkopf2010IGCI,PetersJanzingSchoelkopf2017} advocate that, in the correct causal direction, a short, stable two-part description of the joint distribution exists in which the distribution of the cause and the conditional of the effect given the cause admit largely independent descriptions. This perspective is developed by Janzing, Schölkopf, Peters, and collaborators, spanning information-geometric causal inference \citep{Daniusis2012IGCI} and MDL-based surrogates for Kolmogorov complexity, and is synthesized in Elements of Causal Inference \citep{PetersJanzingSchoelkopf2017}. On the MDL side, Grünwald and others develop stochastic complexity and normalized maximum likelihood \citep{Grunwald2007MDL,Rissanen2007ICS,Shtarkov1987NML} as principled penalties that implement Occam’s razor for model selection and have been used in causal discovery and structure learning. These approaches operate within a representational setting that presupposes code families or model classes and measure simplicity by code length.

The constructor theory of information, developed by Deutsch and Marletto \citep{deutsch2013constructor,Marletto2015CTInformation}, recasts information in terms of possible and impossible tasks on substrates, emphasizing counterfactual laws that govern copying, computation, and communication. This program grounds information in physics yet does not supply a code-length criterion. To our knowledge there is no published account that replaces minimum description length with a constructor-theoretic task cost, recasts Shannon, Kolmogorov, and von Neumann quantities as contextual performance metrics, and applies the resulting principle to causal directionality and robustness across environments. The CT-TDL formulation above is intended to fill that gap by relocating parsimony from syntactic codes to physically permitted constructions and by using robust task cost as the criterion for selecting mechanisms and directions in causal models. Practical surrogates would need to be developed.

\end{document}